\documentclass[conference]{IEEEtran}
\IEEEoverridecommandlockouts
\usepackage{cite}
\usepackage{amsmath,amssymb,amsfonts}
\usepackage{algorithmic}
\usepackage{graphicx}
\usepackage{textcomp}
\usepackage{xcolor}
\def\BibTeX{{\rm B\kern-.05em{\sc i\kern-.025em b}\kern-.08em
    T\kern-.1667em\lower.7ex\hbox{E}\kern-.125emX}}

\usepackage{booktabs}

\usepackage[xindy,style=long,toc=true,nonumberlist,acronym,acronymlists={abbreviations},hyperfirst=true]{glossaries}
\usepackage{algorithm} 
\usepackage{mathtools} 
\mathtoolsset{showonlyrefs=true}
\usepackage{gensymb}
\usepackage{multirow}
\usepackage[caption=false,font=footnotesize]{subfig}
\usepackage{comment}
\usepackage{siunitx}

\newcommand{\PlantForest}{\text{\texttt{PlantForest}}}
\newcommand{\ValidateForest}{\text{\texttt{ValidateForest}}}
\newcommand{\ClimbForest}{\text{\texttt{ClimbForest}}}
\newcommand{\AnalyseForest}{\text{\texttt{AnalyseForest}}}
\newcommand{\SummariseForest}{\text{\texttt{SummariseForest}}}
\newcommand{\ExplainForest}{\text{\texttt{ExplainForest}}}

\newcommand{\TC}{\mathrm{TC}}
\newcommand{\FC}{\mathrm{FC}}

\newcommand{\TS}{\mathrm{TS}}
\newcommand{\TP}{\mathrm{TP}}
\newcommand{\FN}{\mathrm{FN}}
\newcommand{\FP}{\mathrm{FP}}
\newcommand{\TN}{\mathrm{TN}}
\newcommand{\TPR}{\mathrm{TPR}}
\newcommand{\TNR}{\mathrm{TNR}}
\newcommand{\DS}{\mathrm{DS}}

\newcommand{\DEG}{\mathrm{DEG}}
\newcommand{\STA}{\mathrm{STA}}
\newcommand{\IDX}{\mathrm{IDX}}

\newcommand{\DIR}{\mathrm{DIR}}
\newcommand{\DIS}{\mathrm{DIS}}
\newcommand{\DUF}{\mathrm{DUF}}
\newcommand{\DOS}{\mathrm{DOS}}
\newcommand{\DER}{\mathrm{DER}}

\newacronym{ML}{ML}{machine learning}
\newacronym{DL}{DL}{deep learning}
\newacronym{IoU}{IoU}{intersection over union}
\newacronym{SVM}{SVM}{support vector machine}
\newacronym{XAI}{XAI}{explainable AI}
\newacronym{DT}{DT}{decision tree}
\newacronym{AI}{AI}{artificial intelligence}
\newacronym{CNN}{CNN}{convolutional neural network}
\newacronym{DC}{DefChar}{defect characteristic}
\newacronym{RGB}{RGB}{red, green and blue}
\newacronym{HSV}{HSV}{hue, saturation, and brightness}
\newacronym{TPR}{TPR}{true positive rate}
\newacronym{TNR}{TNR}{true negative rate}
\newacronym{RCNN}{R-CNN}{region-based convolutional neural network}
\newacronym{IEMRCNN}{IE-MRCNN}{image-enhanced mask R-CNN}
\newacronym{SHAP}{SHAP}{shapley additive explanations}

\begin{document}
\title{Morphological Image Analysis and Feature Extraction for Reasoning with AI-based Defect Detection and Classification Models}

\author{\IEEEauthorblockN{Jiajun Zhang}
\IEEEauthorblockA{\textit{Dept of Computer Science, School of Science} \\
\textit{Loughborough University, UK}\\
Email: j.zhang8@lboro.ac.uk}
\and
\IEEEauthorblockN{Georgina Cosma\IEEEauthorrefmark{1}}
\IEEEauthorblockA{\textit{Dept of Computer Science, School of Science} \\
\textit{Loughborough University, UK}\\
Email: g.cosma@lboro.ac.uk\\
\IEEEauthorrefmark{1} Corresponding author}
\and
\IEEEauthorblockN{Sarah Bugby}
\IEEEauthorblockA{\textit{Dept of Physics, School of Science} \\
\textit{Loughborough University, UK}\\
Email: s.bugby@lboro.ac.uk}
\and
\IEEEauthorblockN{Axel Finke}
\IEEEauthorblockA{\textit{Dept of Mathematical Sciences, School of Science} \\
\textit{Loughborough University, UK}\\
Email: a.finke@lboro.ac.uk}
\and
\IEEEauthorblockN{Jason Watkins}
\IEEEauthorblockA{\textit{Railston \& Co. Ltd.}\\ 
\textit{Nottingham, UK}\\
Email: jason@railstons.com}
\thanks{This research is funded through joint funding by the School of Science at Loughborough University with industrial support from Railston \& Co Ltd.}}

\maketitle

\begin{abstract}
As the use of \gls{AI} models becomes more prevalent in industries such as engineering and manufacturing, it is essential that these models provide transparent reasoning behind their predictions. This paper proposes the AI-Reasoner, which extracts the morphological characteristics of defects (DefChars) from images and utilises decision trees to reason with the DefChar values. Thereafter, the AI-Reasoner exports visualisations (i.e.\ charts) and textual explanations to provide insights into outputs made by masked-based defect detection and classification models. It also provides effective mitigation strategies to enhance data pre-processing and overall model performance. The AI-Reasoner was tested on explaining the outputs of an IE Mask R-CNN model using a set of 366 images containing defects. The results demonstrated its effectiveness in explaining the IE Mask R-CNN model's predictions. Overall, the proposed AI-Reasoner provides a solution for improving the performance of \gls{AI} models in industrial applications that require defect analysis.
\end{abstract}


\glsreset{AI}
\section{Introduction}
\label{sec:introduction}

\begin{figure*}[!t]
\centering
\includegraphics[width=\linewidth]{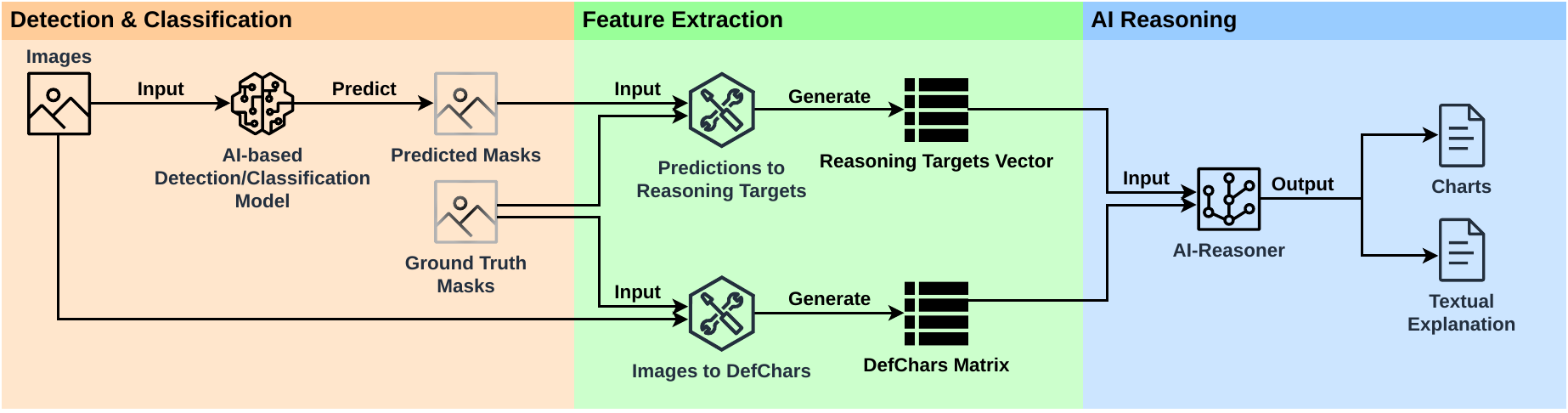}
    \caption{AI-Reasoner architecture.}
    \label{fig:framework}
\end{figure*}

As \textit{\gls{AI}} models become increasingly prevalent in industries such as manufacturing, there is a growing need to explain the reasoning behind \textit{\gls{ML}} model outputs. Developing a comprehensive understanding of \gls{ML} model outputs empowers end-users to make informed decisions based on the results \cite{Ahmedetal2022}. Moreover, this understanding equips developers with vital information to effectively address and overcome any limitations inherent in the model.
To address this challenge, researchers have developed a field of study known as \textit{\gls{XAI}}. \gls{XAI} techniques use various strategies to provide interpretations for the \gls{AI} model and its outputs, allowing non-experts to understand how the models arrive at their predictions. These strategies may involve generating explanations that can be understood by humans, visualising the model's decision-making process, or identifying the key features that the model relies on to make its predictions. By providing explanations of an \gls{AI} model's outputs, \gls{XAI} can help to increase the transparency and trustworthiness of the model and enable non-experts to make informed decisions based on the model's predictions. This can ultimately lead to improved safety, quality, and efficiency in industrial applications.

In the application of \gls{ML} to defect detection \cite{01.da0,01.da1,01.da2,01.da3,01.da4} and analysis \cite{01.da10,01.da11,01.da12}, there have been some attempts to provide explainability of outputs. Fanci \textit{et al.\@} \cite{01.dc_rw1} used image morphology analysis and human visual characteristics (i.e.\ area, roundness, ratio, brightness, and texture) to identify image defects, and demonstrated that their method could successfully distinguish defects. Lafor and Peansupap \cite{01.dc_rw2} proposed a defect detection system that quantifies defects into a feature list (i.e.\ region, edges, scale, interest points, and texture) and which achieved \SI{94}{\percent} detection accuracy in a tiling defect detection task, and effectively reduced engineers' subjective judgements of aesthetic faults. Dong \textit{et al.\@} \cite{01.dc_rw3} extracted geometry-, texture- and vision-based features from images to perform weld defect analysis using a support vector machine model, showing that defects could be identified with high accuracy over \SI{90}{\percent} in a pipeline weld defect detection task. Yan and Gao \cite{01.def_fe} extracted a set of optical image features, including colour-, texture- and shape-based features, from steel surface defect images and used those for a classification task for an engineering enterprise, and their model's performance reached an average of \SI{98}{\percent} accuracy.

One approach to \gls{XAI} is through the use of saliency map methods \cite{01.xai_lcd, 01.xai_defect, 01.xai_impdl, 01.xai_shap1, 01.xai_shap2}, that can explain neural network models by highlighting interesting regions in a defect image.
Another approach is through the use of simplification and feature-relevance methods \cite{01.xai_fr1, 01.xai_fr2, 01.xai_fr3} that can provide \gls{ML} model explanations by extracting and analysing defect features from images and tabular data. These methods can help identify the most important features that a model relies on to make its predictions, enabling non-experts to better understand how the model derived its decisions. 

This paper proposes the AI-Reasoner that facilitates reasoning with \gls{AI} outputs. The contributions are:
\begin{itemize}
    \item A new set of \textit{\glspl{DC}} for describing defects based on their morphological characteristics. \gls{DC} values represent quantitative information about defects, including their colour, shape, and meta aspects. \glspl{DC} can be utilised in \gls{AI}-based models for reasoning and defect analysis tasks. 
    \item A novel AI-Reasoner that extracts \glspl{DC} from images and generates a set of charts and textual descriptions to provide visualisation and reasoning with \gls{AI} outputs.
    \item Empirical demonstration of the AI-Reasoner on the outputs of a Mask \textit{\gls{RCNN}} model using a set of 366 images containing defects. The results demonstrate the usefulness of the proposed \glspl{DC} and AI-Reasoner in explaining the model's predictions. The AI-Reasoner offers effective data pre-processing mitigation strategies aimed at enhancing model performance, and which can also save experimental time. Furthermore, this paper discusses the suitability of \gls{SHAP} for Reasoning with Defect Predictions.
\end{itemize}

\section{Proposed AI-Reasoner}

This section presents the AI-Reasoner's components, as illustrated in Fig.~\ref{fig:framework}.

\subsection{Detection and Classification}
A mask-based \textit{\gls{DL}} model can be utilised to detect or classify defect regions. Such a model can be a Mask \gls{RCNN} that takes images and predicts the presence or categories of the defects. 

\subsection{Feature Extraction: Predictions to Reasoning Targets}
\label{sec:create_rt}

The AI-Reasoner is able to reason the AI predictions for a detection task or a classification task. Let $X$ be an image. Assume that this image shows $I$ defects in pairwise disjoint (i.e.\ non-overlapping) regions. In other words, $X \setminus (X_1 \cup  \dotsb \cup X_I)$ is the remaining part of the image which does not contain any defects. Furthermore, let $L_i = L(X_i) \in \{1,\dotsc,K\}$ denote the type (category) of the defect in the $i$th region.

An \gls{AI} model for defect detection will then output a prediction consisting of $J$ pairs $\smash{(\widehat{X}_1, \widehat{L}_1), \dotsc, (\widehat{X}_J, \widehat{L}_J)}$, where $\smash{\widehat{X}_j \subseteq X}$ is a predicted defect region and $\smash{\widehat{L}_j = \widehat{L}(\widehat{X}_j) \in \{1, \dotsc, K\}}$ is the predicted type of defect in that region. 

Throughout this work, it is assumed that a predicted defect can always be matched to at most one true defect; and that a true defect can always be matched to at most one predicted defect, e.g., via the \textit{\gls{IoU}} value.

For the remainder of this work, reasoning targets (i.e.\ true positives and false negatives) will be defined as follows.
\begin{itemize}
    \item \textbf{Detection.} If only \textit{detection} (but not classification) of defects is of interest, one can consider which of the true defects have been detected (or not) and set $C = (C_1, \dotsc, C_I)$ and $D = (D_1, \dotsc, D_I)$, where
    \begin{align}
    \label{eq:detection:option:1}
      C_i & =
      \begin{cases}
        1, & \text{if $X_i$ matches some $\widehat{X}_j$},\\ 
        0, & \text{otherwise,}
      \end{cases}\\
      D_i &=
      \begin{cases}
        1, & \text{if $X_i$ does not match any $\widehat{X}_j$},\\ 
        0, & \text{otherwise,}
      \end{cases} 
      \label{eq:detection:option:2}
    \end{align}
    for $i = 1,\dotsc, I$.
    \item \textbf{Classification.} If only \textit{classification} (of already known/correctly detected) defects is of interest (in which case $I = J$), one can set $C' = (C_1', \dotsc, C_I')$ and $D' = (D_1', \dotsc, D_I')$, where
    \begin{align}
      C_i' =
      \begin{cases}
        1, & \text{if $\widehat{L}_i = L_i$},\\ 
        0, & \text{otherwise,}
      \end{cases}
      \quad
      D_i' =
      \begin{cases}
        1, & \text{if $\widehat{L}_i \neq L_i$},\\ 
        0, & \text{otherwise,}
      \end{cases} 
    \end{align}
    for $i = 1,\dotsc, I$.

    \item \textbf{Joint detection and classification.} If both \textit{detection} and \textit{classification} of defects are of interest, one can specify $C_i$ and $D_i$ as in (\refeq{eq:detection:option:1}) and (\refeq{eq:detection:option:2}) and then set
    \begin{align}
      C_i' &=
      \begin{cases}
        1, & \text{if $C_i = 1$ and $\widehat{L}(X_i) = L_i$},\\ 
        0, & \text{otherwise,}
      \end{cases}\\
      D_i' &=
      \begin{cases}
        1, & \text{if $C_i = 1$ and $\widehat{L}(X_i) \neq L_i$},\\ 
        0, & \text{otherwise,}
      \end{cases}
    \end{align}
    for $i = 1,\dotsc, I$, where $\widehat{L}(X_i)$ is the predicted category of the $i$th true defect. Finally, set $C = (C_1, \dotsc, C_I)$, $C' = (C_1', \dotsc, C_I')$, $D = (D_1, \dotsc, D_I)$, $D' = (D_1', \dotsc, D_I')$.
\end{itemize}

\subsection{Feature Extraction: Images to \gls{DC}} \label{sec:dc}

\begin{table*}[ht]
\caption{Proposed \glspl{DC}.} 
\centering
\renewcommand{\arraystretch}{1.3}
\begin{tabular}{lcl}
\toprule
 \multicolumn{3}{c}{\textbf{Colour Information extracted and stored separately for the defect and background areas}} \\ \midrule
 \textbf{DefChar Name} & \textbf{Value Range} & \textbf{Description} \\ \midrule
 Average Hue & $\{0,1,\dotsc, 359\}$ & \parbox{0.63\linewidth}{Average hue value} \\ 
 Mode of Hue & $\{0,1,\dotsc, 359\}$ & \parbox{0.63\linewidth}{Most frequent hue value} \\ 
 Unique number of Hue values & $\{1,2,\dotsc, 360\}$ & \parbox{0.63\linewidth}{Number of unique hue values} \\ 
 Hue Range & $\{0,1,\dotsc, 180\}$ &  \parbox{0.63\linewidth}{Difference of maximum and minimum hue value } \\ 
 Average Saturation & $\{0,1,\dotsc, 254\}$ & \parbox{0.63\linewidth}{Average saturation value} \\ 
 Mode of Saturation & $\{0,1,\dotsc, 254\}$ & \parbox{0.63\linewidth}{Most frequent saturation value} \\ 
 Unique number of Saturation & $\{1,2,\dotsc, 255\}$ & \parbox{0.63\linewidth}{Number of unique saturation values}\\ 
 Saturation Range & $\{0,1,\dotsc, 254\}$ & \parbox{0.63\linewidth}{Difference of maximum and minimum saturation values} \\ 
 Average Brightness & $\{0,1,\dotsc, 254\}$ & \parbox{0.63\linewidth}{Average brightness value} \\ 
 Mode of Brightness & $\{0,1,\dotsc, 254\}$ & \parbox{0.63\linewidth}{Most frequent brightness value} \\ 
 Unique number of Brightness & $\{1,2,\dotsc, 255\}$ & \parbox{0.63\linewidth}{Unique brightness values} \\ 
 Brightness Range & $\{0,1,\dotsc, 254\}$ & \parbox{0.63\linewidth}{Difference of maximum and minimum brightness value} \\ \midrule
 \multicolumn{3}{c}{\textbf{Colour Complexity}} \\ \midrule
 \textbf{DefChar Name} & \textbf{Value Range} & \textbf{Description} \\ \midrule
 Hue Difference & $[0,1]$ & \parbox{0.63\linewidth}{Hue frequency distribution difference between the defect and background areas} \\ 
 Saturation Difference & $[0,1]$ & \parbox{0.63\linewidth}{Saturation frequency distribution difference between the defect and background areas} \\
 Brightness Difference & $[0,1]$ & \parbox{0.63\linewidth}{Brightness frequency distribution difference between the defect and background areas} \\ \midrule
 \multicolumn{3}{c}{\textbf{Shape Information}} \\ \midrule
 \textbf{DefChar Name} & \textbf{Value Range} & \textbf{Description} \\ \midrule
 Number of Edges & $\{3,4,\dotsc\}$ & \parbox{0.63\linewidth}{Number of edges of the defect polygon areas} \\
 Coverage & $[0,1]$ & \parbox{0.63\linewidth}{Percentage of the defect polygon area covered by its bounding box} \\ 
 Aspect Ratio & $[0,1]$ & \parbox{0.63\linewidth}{Ratio between the width and height of defect bounding box} \\ 
 Average Turning Angles & $\{1,2,\dotsc, 180\}$ & \parbox{0.63\linewidth}{Average value of vertex angles of the defect polygon area} \\ 
 Mode of Turning Angle & $\{1,2, \dotsc, 180\}$ & \parbox{0.63\linewidth}{Value of vertex angles that appears the most often in the defect polygon} \\  \midrule
 \multicolumn{3}{c}{\textbf{Shape Complexity}} \\ \midrule
 \textbf{DefChar Name} & \textbf{Value Range} & \textbf{Description} \\ \midrule
 Edge Ratio & $[0,1]$ & \parbox{0.63\linewidth}{Average length ratio between two adjacent edges in the defect polygon area} \\ 
 Followed Turns & $[0,1]$ & \parbox{0.63\linewidth}{Proportion of two adjacent vertices which turn to the same direction in the defect polygon area} \\ 
Small Turns & $[0,1]$ & \parbox{0.63\linewidth}{Percentage of vertices which are smaller than 90$\degree$ in the defect polygon area} \\
 Reversed Turns & $[0,1]$ & \parbox{0.63\linewidth}{Proportion of two adjacent vertices which turn to a different direction in the defect polygon area} \\ \midrule
 \multicolumn{3}{c}{\textbf{Meta Information}} \\ \midrule
 \textbf{DefChar Name} & \textbf{Value Range} & \textbf{Description} \\ \midrule
 Defect Size & $\{1,2,\dotsc\}$ & \parbox{0.63\linewidth}{Number of pixels in the defect polygon area} \\
 Neighbour Distance & $\{0,1,2\}$ & \parbox{0.63\linewidth}{Categorised distances to the nearest neighbour, 0$\rightarrow$Short ($\leq$\SI{100}{px}); 1$\rightarrow$Long; 2$\rightarrow$No Neighbour.} \\ \bottomrule
\end{tabular}
\label{tab:defchar}
\end{table*}
\glspl{DC} are a new set of defect characteristics that are extracted from images by analysing the \textit{\gls{RGB}} values and the polygon-shaped defect regions/masks in an image. Table~\ref{tab:defchar} presents the complete set of \glspl{DC}.

\noindent \textbf{Color characteristics:} extract colour information using the \textit{\gls{HSV}} values, converted from \gls{RGB} values. \gls{HSV} values can provide more intuitive colour properties than \gls{RGB} values. The colour complexity features capture the frequency distribution differences of the \gls{HSV} values between the defect and background areas.\\
\textbf{Shape characteristics:} extract shape information from polygon annotations, including bounding boxes, vertices and edges. Shape complexity indicates the shape irregularity of the defect by calculating four values: the edge ratio, follow turning, reverse turning and small turning.\\
\textbf{Meta characteristics:} `Defect size' provides an indication of the severity of the defect. `Distance to the nearest neighbour defect' provides information about a defect's environment.\\

\subsection{AI Reasoning}
\label{sec:cs1.dtae}
The AI-Reasoner uses an ensemble \textit{\gls{DT}} to reason with the outputs of an \gls{AI}-based defect detection model. Algorithm~\ref{code:r_e} introduces the pseudocode for the proposed AI-Reasoner, which comprises: $\PlantForest$, $\ValidateForest$, $\ClimbForest$, $\AnalyseForest$,  $\SummariseForest$ and $\ExplainForest$. 

\begin{algorithm}[ht]
\caption{AI-Reasoner.}
\hspace*{0.01in} {\bf Input:} 
 DefChar matrix $E$ of size $m \times 38$, where $m$ represents the number of defects in the dataset.\\
\hspace*{0.43in} Reasoning target vector $T = C$ or $T = D$ or $T = C'$ or $T = D'$, where $C$, $D$, $C'$ and $D'$ are as in Section~\ref{sec:create_rt}\\ 
\hspace*{0.01in} {\bf Output:} Reasoning result $R$ (charts and descriptions) 
\label{code:r_e}
    \begin{algorithmic}[1]
        \STATE $M \leftarrow \PlantForest(E, T)$ 
        \STATE $V \leftarrow \ValidateForest(M, E, T)$
        \PRINT $V$ \% defects have been correctly reasoned
        \STATE $O \leftarrow \ClimbForest(M)$ 
        \STATE $A \leftarrow \AnalyseForest(O)$
        \STATE $S \leftarrow \SummariseForest(A)$ 
        \STATE $R \leftarrow \ExplainForest(S)$

        \RETURN $R$ 
    \end{algorithmic}
\end{algorithm}
Empirical evaluations were carried out to determine the optimal parameters for the ensemble \glspl{DT}. These are the recommended parameters when using the proposed AI-Reasoner. \\
$\PlantForest$ takes the \gls{DC} matrix $E$ and reasoning target vector $T$ as inputs where $T$ is the reasoning target $C$, $D$, $C'$ or $D'$ (described in Section~\ref{sec:create_rt}). Then, 200 \glspl{DT} are created with the untrimmed setting (i.e.\ \texttt{max depth = -1}: unrestricted tree depth, \texttt{min split = 2}: at least two samples in a node when splitting it, and \texttt{min leaf = 1}: at least one sample contained in a leaf node) for exploring reasons behind the predictions of an AI-based defect detection model. All trained \glspl{DT} are stored in a list $M$ for the next step. \\
$\ValidateForest$ evaluates the learning performance of each trained \gls{DT} model $M$ using the metrics (\textit{\gls{TPR}} and \textit{\gls{TNR}}) defined in Section~\ref{sec:cs1.mve}. The input data (\gls{DC} matrix $E$ and reasoning target vector $T$) are utilised to assess how many defects were correctly learnt by each trained \gls{DT}. Then, a learning score $V$ provides the overall learning performance of the AI-Reasoner and the equation is shown in (\refeq{eq:validscore}).\\
$\ClimbForest$ parses all trained \glspl{DT} and recursively reads the nodes in each tree and extracts tree information (i.e.\ \gls{DC} features, split condition, the sample distributions in the node and its child nodes). The extracted information of every node is stored in a list $O$ for further steps.\\
$\AnalyseForest$ analyses the extracted list of \glspl{DT}' nodes $O$ and computes a set of values to determine the importance of each node in each \gls{DT}. A \textit{decision score}, $\DS$, is the average split ratio in which the parent node divides the samples into its two child nodes:
\begin{equation}
    \label{eq:ds}
    \DS = \frac{1}{2} \biggl(\biggl\lvert \frac{\TC_1 - \FC_1}{N_1} \biggr\rvert + \biggl\lvert \frac{\TC_0 - \FC_0}{N_0} \biggr\rvert \biggr),
\end{equation}
where
\begin{itemize}
    \item $N_1$ is the number of samples whose reasoning target values are 1 in a node;
    \item $N_0$ is the number of samples whose reasoning target values are 0 in a node;
    \item $\TC_1$ is the number of samples whose reasoning target values are 1 in its true child node;
    \item $\TC_0$ is the number of samples whose reasoning target values are 0 in its true child node;
    \item $\FC_1$ is the number of samples whose reasoning target values are 1 in its false child node;
    \item $\FC_0$ is the number of samples whose reasoning target values are 0 in its false child node.
    \label{list:sd}
\end{itemize}
Furthermore, the \textit{distinguish score}, $\TS$, measures the percentage in which the parent node isolates the samples in its two child nodes:
\begin{equation}
    \label{eq:ts}
    \TS = \biggl\lvert \frac{\TC_1}{N_1} - \frac{\TC_0}{N_0} \biggr\rvert
    = \biggl\lvert \frac{\FC_1}{N_1} - \frac{\FC_0}{N_0} \biggr\rvert.
\end{equation}
Finally, \textit{usage of samples}, $U$, is the percentage of samples contained in a node:
\begin{equation}
    \label{eq:u}
    U = \frac{N_1+N_0}{N},
\end{equation}
where $N$ is the total number of samples, i.e.\ at the root node.

\begin{table*}[ht]
\caption{Information about a Node's Decision Degree, Status and Bonus.}
\renewcommand{\arraystretch}{1.3}
\centering
\begin{tabular}{lcclcc}
\toprule
\multicolumn{3}{c}{$\mathbf{DEG}$} & \multicolumn{3}{c}{$\mathbf{STA}$} \\ \cmidrule(r){1-3}\cmidrule(l){4-6}
\textbf{$\mathbf{TS}$ threshold} &\textbf{Degree} & $\mathbf{B_{\mathbf{deg}}}$ & \textbf{Condition} & \textbf{Status} & $\mathbf{B_{\mathbf{sta}}}$ \\ \midrule
$\TS = 0$ & empty & 0 & $\TC_1 = 0 \text{ or } \TC_0 = 0$ & confirmation & 0.5 \\
$0 < \TS < 0.25$ & weak & 0.1 & $\TC_1 = \FC_1 \text{ or } \TC_0 = \FC_0$ & half reduction & 0.2 \\
$0.25 \leq \TS < 0.5$ & middle & 0.2 & all other conditions & reduction & 0  \\
$0.5 \leq \TS < 1$ & strong & 0.3 &  \\ 
$\TS = 1$ & full & 0.5 &  \\ \bottomrule
\end{tabular}
\label{tab:dds}
\end{table*}
A set of additional values are computed. The \textit{node importance index}, $\IDX$, is calculated to state a node's capability in distinguishing samples that have different reasoning target values:
\begin{equation}
    \label{eq:idx}
    \IDX = U \times ((1+\DS) \times \TS + B_{\mathrm{deg}} + B_{\mathrm{sta}}).
\end{equation}
Decision degree bonus ($B_{\mathrm{deg}}$) and decision status bonus ($B_{\mathrm{sta}}$) are calculated based on decision degree ($\DEG$) and decision status ($\STA$) (shown in Table~\ref{tab:dds}), and these two values are used to amplify the $\IDX$ value. $\DEG$ expresses the sample isolation degree (i.e.\ empty, weak, middle, strong and full) according to $\TS$. $\STA$ expresses the node's splitting status (i.e.\ confirmation, half reduction, and reduction) according to the values of $\TC_1$, $\TC_0$, $\FC_1$, $\FC_0$.
Decision direction ($\DIR$) is a boolean signal that indicates which child node contains more samples that reasoning target values are 1. The output is an extended list $A$, which contains the list $O$ and these analysed values. \\
$\SummariseForest$ computes the importance of each \gls{DC} by summarising the list $A$. Firstly, all analysed nodes are divided into groups by \gls{DC}; hence 38 groups of nodes are constructed where each node group corresponds to a unique \gls{DC}. Next, DefChar Importance Index ($\DIS$), DefChar Usage Frequency ($\DUF$), DefChar Overall Score ($\DOS$), DefChar Effective Range ($\DER$) are calculated to quantify the importance of each \gls{DC} as follows.
\begin{itemize}
    \item $\DIS$ is calculated by averaging the node importance indices $\IDX$ in the grouped nodes; a higher value indicates that the \gls{DC} can easily affect the \gls{AI} model's detection/classification performance. 
    \item $\DUF$ is calculated by averaging the occurrence of the \gls{DC} used in the ensemble \gls{DT}; a higher value implies that the \gls{DC} is essential to split the reasoning targets. 
    \item $\DOS$ is the overall score for each \gls{DC} calculated by multiplying the $\DIS$ and $\DUF$ with a 3:1 ratio. 
    \item $\DER$ is a value range to show the \gls{DC} value interval that can affect the \gls{AI} model's predictions.
\end{itemize}
The metrics defined in this section are utilised by $\ExplainForest$ to create the AI-Reasoner's output (i.e.\ charts and textual descriptions) for enabling the end-user to reason with a model's outputs.
$\ExplainForest$ visualises the array-format reasoning result $S$ using charts and provides textual descriptions that include mitigation strategies $R$ to help users gain reasons behind the \gls{AI} results and to take steps in improving the dataset that was used to train the models. Sample outputs are shown in the paper's Github repository\footnote{https://github.com/edgetrier/AI-Reasoner}. 

\subsection{Metrics for Evaluating the AI-Reasoner's Learning}
\label{sec:cs1.mve}
\gls{TPR} and \gls{TNR} evaluate the learning performance of each \gls{DT} and are defined as
\begin{align}
\TPR = \frac{\TP}{\TP + \FN}, \quad 
\TNR = \frac{\TN}{\TN + \FP}, \label{eq:tnr}
\end{align}
where
\begin{itemize}
\item $\TP$ is the total number of defects whose reasoning target values are 1 and were correctly learnt by the \gls{DT};
\item $\TN$ is the total number of defects whose reasoning target values are 0 and were correctly learnt by the \gls{DT};
\item $\FP$ is the total number of defects whose reasoning target values are 1 and were not correctly learnt by \gls{DT};
\item $\FN$ is the total number of defects whose reasoning target values are 0 and were not correctly learnt by \gls{DT}. 
\end{itemize}
\gls{TPR} measures the ability of a model to correctly identify positive instances. \gls{TNR} measures the ability of a model to correctly identify negative instances. 

The learning score evaluates the overall learning performance of the reasoning model:
\begin{equation}
\text{Learning Score} = \frac{1}{n} \sum^{n}_{i=1} \frac{\TPR_{i} + \TNR_{i}}{2},
\label{eq:validscore}
\end{equation}
where $n$ is the number (i.e.\ $n=200$) of \glspl{DT} that comprise the reasoning model; $i$ is the index of a \gls{DT}; $\TPR_{i}$ and $\TNR_{i}$ are the TPR and the TNR of the $i$th \gls{DT}.

\section{Experiment Methodology -- Utilising \glspl{DC} to reason  with \gls{AI} models}
\label{sec:cs1_AIR}

\begin{table*}[ht]
\centering
\caption{Overview of Prediction Results from Image-enhanced Mask R-CNN \cite{01.IEMRCNN}.}
\renewcommand{\arraystretch}{1.3}
\begin{tabular}{lccc}
\toprule
\multirow{2}{*}{\textbf{Prediction Results}} & \multicolumn{3}{c}{\textbf{Number (Percentage) of Detected Defects}} \\ \cmidrule{2-4}
& \textbf{Augmented} & \textbf{Greyscaled} & \textbf{Image-Enhanced} \\ \midrule
Detected $C$ & 311 (\SI{84.97}{\percent}) & 298 (\SI{81.42}{\percent}) & 308 (\SI{84.15}{\percent}) \\ 
Undetected $D$ & 55 (\SI{15.03}{\percent}) & 68 (\SI{18.58}{\percent}) & 58 (\SI{15.85}{\percent}) \\ 
Correctly Classified $C'$ & 296 (\SI{80.87}{\percent}) & 287 (\SI{78.42}{\percent}) & 295 (\SI{80.60}{\percent}) \\
Misclassified $D'$ & 15 (\SI{4.10}{\percent}) & 11 (\SI{3.00}{\percent}) & 13  (\SI{3.55}{\percent}) \\ 
Total & \multicolumn{3}{c}{366 (\SI{100.00}{\percent})} \\ \bottomrule
\end{tabular}
\label{tab:data_dist}
\end{table*}
 
The proposed AI-Reasoner is applied to reason with the outputs of the \gls{IEMRCNN} model proposed by Zhang \textit{et al.\@} \cite{01.IEMRCNN} which can detect the presence and types of wind turbine blade defects. Their paper evaluates the model's defect detection and type classification performance on an augmented dataset of defects (v1), the augmented dataset greyscaled (v2), and the augmented dataset after image enhancement (but not greyscaled) (v3). Their prediction results are shown in Table~\ref{tab:data_dist}.

\subsection{Dataset}
\label{sec:cs1.dataset}

The dataset utilised for the experiments was provided by Zhang \textit{et al.\@} \cite{01.IEMRCNN}. Each defect is represented as a set of \gls{DC} features stored in a 366 $\times$ 38 matrix $E$. The ground truth label (i.e.\ region and type) of each defect is stored in a set of pairs $(X,L)$ where each pair $(X_i,L_i)$ corresponds to the $i$th row of matrix $E$. The predicted label of each defect is stored in a set of pairs $(\widehat{X},\widehat{L})$.

\subsection{Methodology}
\label{sec:method}

\begin{table}[ht]
\centering
\caption{Combinations of \glspl{DC}.}
\renewcommand{\arraystretch}{1.3}
\begin{tabular}{cccc}
\toprule
\textbf{Combination} & \textbf{Color DefChar} & \textbf{Shape DefChar} & \textbf{Meta DefChar} \\ \midrule
Color & $\checkmark$ &  &  \\ 
Shape &  & $\checkmark$ &  \\ 
Meta &  &  & $\checkmark$ \\ 
Color-Shape & $\checkmark$ & $\checkmark$ & \\
All \glspl{DC} & $\checkmark$ & $\checkmark$ & $\checkmark$ \\ \bottomrule
\end{tabular}

\label{tab:dccomb}
\end{table}

\noindent \textbf{Step 1:} Apply the Mask \gls{RCNN} model to the dataset containing 366 defects with ground truth labels (i.e.\ regions $X$, types $L$) and obtain predicted labels (i.e.\ regions $\widehat{X}$, types $\widehat{L}$) (described in Section~\ref{sec:cs1.dataset}). \\
\textbf{Step 2:} Extract the \gls{DC} matrix $E$ ($366 \times 38$) of the images and rescale the \gls{DC} values using the min-max scaling method (described in Section~\ref{sec:dc}). \\
\textbf{Step 3:} Convert and merge each image's ground truth labels $(X,L)$ and predicted labels $(\widehat{X},\widehat{L})$ of the Mask \gls{RCNN} model to four separate reasoning target vectors $C$, $D$, $C'$ and $D'$. Vectors $C$ and $D$ hold the reasoning targets of the correct and incorrect model outputs of the detection task, and vectors $C'$ and $D'$ hold the reasoning targets of the correct and incorrect outputs of the classification task (see Section~\ref{sec:create_rt}).\\
\textbf{Step 4:} Apply the AI-Reasoner (see Section~\ref{sec:cs1.dtae}) to the \gls{DC} matrix $E$ and each reasoning target vector $T \in \{C, D, C', D'\}$. Conduct four experiments, each with a different reasoning target: 1) reasoning with the outputs that were correctly detected; 2) reasoning with the outputs that were not detected; 3) reasoning with the outputs whose types were correctly classified; and 4) reasoning with the outputs whose types were not correctly classified. \\
\textbf{Step 5:} Evaluate the learning performance of the AI-Reasoner across each experiment using the learning score metric described in Section~\ref{sec:cs1.mve}.
The learning scores are averaged across four reasoning targets and presented in Table~\ref{tab:avgres} that also contains the results when tuning the \gls{DT} with different parameter settings (\texttt{max depth}, \texttt{min split} and \texttt{min leaf}) and using different \gls{DC} combinations (see Table~\ref{tab:dccomb}).\\
\textbf{Step 6:} AI-Reasoner interprets the outputs, presents charts, textual explanations, and suggests mitigation strategies to the user for improving prediction performance. These strategies are presented to the end-user in textual format (see Section~\ref{sec:outint}). The user can follow the proposed mitigation strategies to improve their dataset and the model's performance. 

\section{Results}

\begin{table*}[ht]
\centering
\caption{Average learning scores for each \gls{DC} combination across four reasoning targets (see Section III-B step 4).}
\label{tab:avgres}
\renewcommand{\arraystretch}{1.3}
\begin{tabular}{cccS[table-format=3.2]S[table-format=3.2]S[table-format=3.2]S[table-format=3.2]S[table-format=3.2]}
\toprule
\multicolumn{3}{c}{\textbf{Parameters}} & \multicolumn{5}{c}{\textbf{DefChar Combination}} \\ \cmidrule(r){1-3} \cmidrule(l){4-8}
\textbf{Max Depth} & \textbf{Min Split} & \textbf{Min Leaf} & \textbf{Color} & \textbf{Shape} & \textbf{Meta} & \textbf{Color-Shape} & \textbf{All \glspl{DC}} \\ \midrule
1 & 2 & 1 & \SI{50.23}{\percent} & \SI{50.00}{\percent} & \SI{50.09}{\percent} & \SI{50.20}{\percent} & \textbf{\SI[detect-weight]{50.31}{\percent}} \\ 
5 & 2 & 1 & \SI{68.75}{\percent} & \SI{55.69}{\percent} & \SI{52.24}{\percent} & \SI{69.52}{\percent} & \textbf{\SI[detect-weight]{71.88}{\percent}} \\
10 & 2 & 1 & \SI{91.67}{\percent} & \SI{75.57}{\percent} & \SI{71.52}{\percent} & \SI{93.14}{\percent} & \textbf{\SI[detect-weight]{94.72}{\percent}} \\ 
Infinity & 2 & 1 & \SI{100.00}{\percent} & \SI{96.60}{\percent} & \SI{96.60}{\percent} & \SI{100.00}{\percent} & \textbf{\SI[detect-weight]{100.00}{\percent}} \\
Infinity & 2 & 3 & \SI{79.24}{\percent} & \SI{67.48}{\percent} & \SI{60.33}{\percent} & \SI{81.02}{\percent} & \textbf{\SI[detect-weight]{84.62}{\percent}} \\  
Infinity & 2 & 5 & \SI{69.16}{\percent} & \SI{56.61}{\percent} & \SI{53.54}{\percent} & \SI{70.57}{\percent} & \textbf{\SI[detect-weight]{75.88}{\percent}} \\  
Infinity & 5 & 1 & \SI{88.84}{\percent} & \SI{79.45}{\percent} & \SI{74.61}{\percent} & \SI{89.06}{\percent} & \textbf{\SI[detect-weight]{91.28}{\percent}} \\ 
Infinity & 5 & 3 & \SI{79.26}{\percent} & \SI{67.52}{\percent} & \SI{60.37}{\percent} & \SI{80.87}{\percent} & \textbf{\SI[detect-weight]{84.55}{\percent}} \\ \bottomrule
\end{tabular}
\end{table*}
\subsection{Reasoning Performance when using \gls{DC}s}

This section describes the learning performance of the AI-Reasoner when using different parameters and combinations of \glspl{DC}. The learning scores for each parameter setting and combination are computed by averaging the learning scores across the four reasoning targets (see Section~\ref{sec:method} step 4) and are shown in Table~\ref{tab:avgres}. The highest learning score of each model is marked in bold text.
The AI-Reasoner achieves a higher learning score if the \gls{DT} is set with a deep tree depth, a small number of splits, and a small number of leaves. The learning score gradually increased from \SI{50.31}{\percent} (\texttt{max depth = 1}, \texttt{min split = 2}, and \texttt{min leaf = 1}) to \SI{100}{\percent} (\texttt{max depth = infinity}, \texttt{min split = 2}, and \texttt{min leaf = 1}) when applying all \glspl{DC}. The AI-Reasoner achieved the highest learning score of \SI{100}{\percent} when using all \glspl{DC} and the \gls{DT} setting (\texttt{max depth = infinity}, \texttt{min split = 2}, and \texttt{min leaf = 1}); hence, this DefChar combination and \gls{DT} settings are applied to the \glspl{DT} that comprise AI-Reasoner.

\subsection{Interpretation of the AI-Reasoner's Outputs}
\label{sec:outint}

\begin{figure*}[!ht]
    \centering
    \subfloat[Undetected Cases $D$.]{\includegraphics[width=3in]{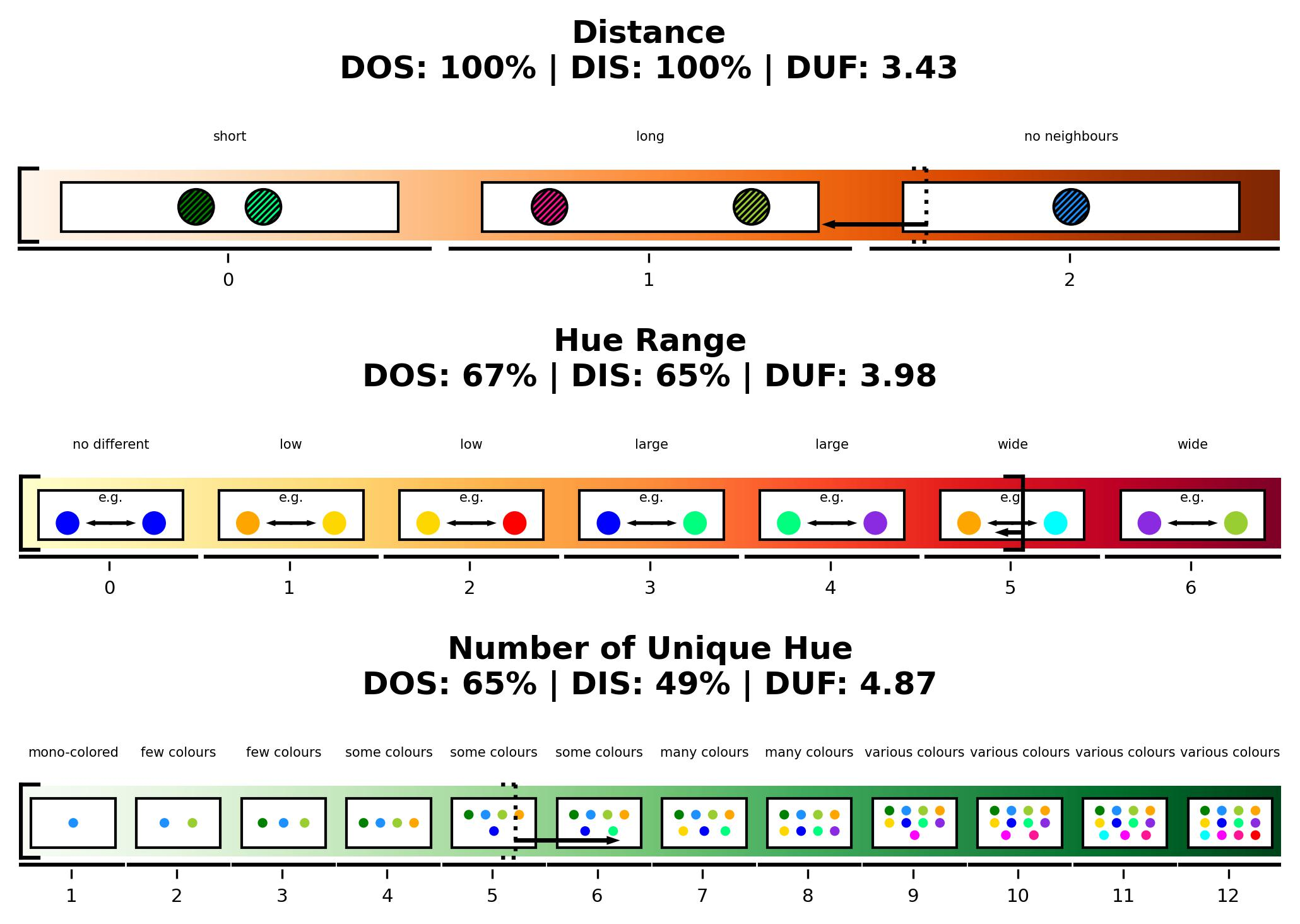}}
    \label{fig:det}
    \hfil
    \subfloat[Misclassified Cases $D'$.]{\includegraphics[width=3in]{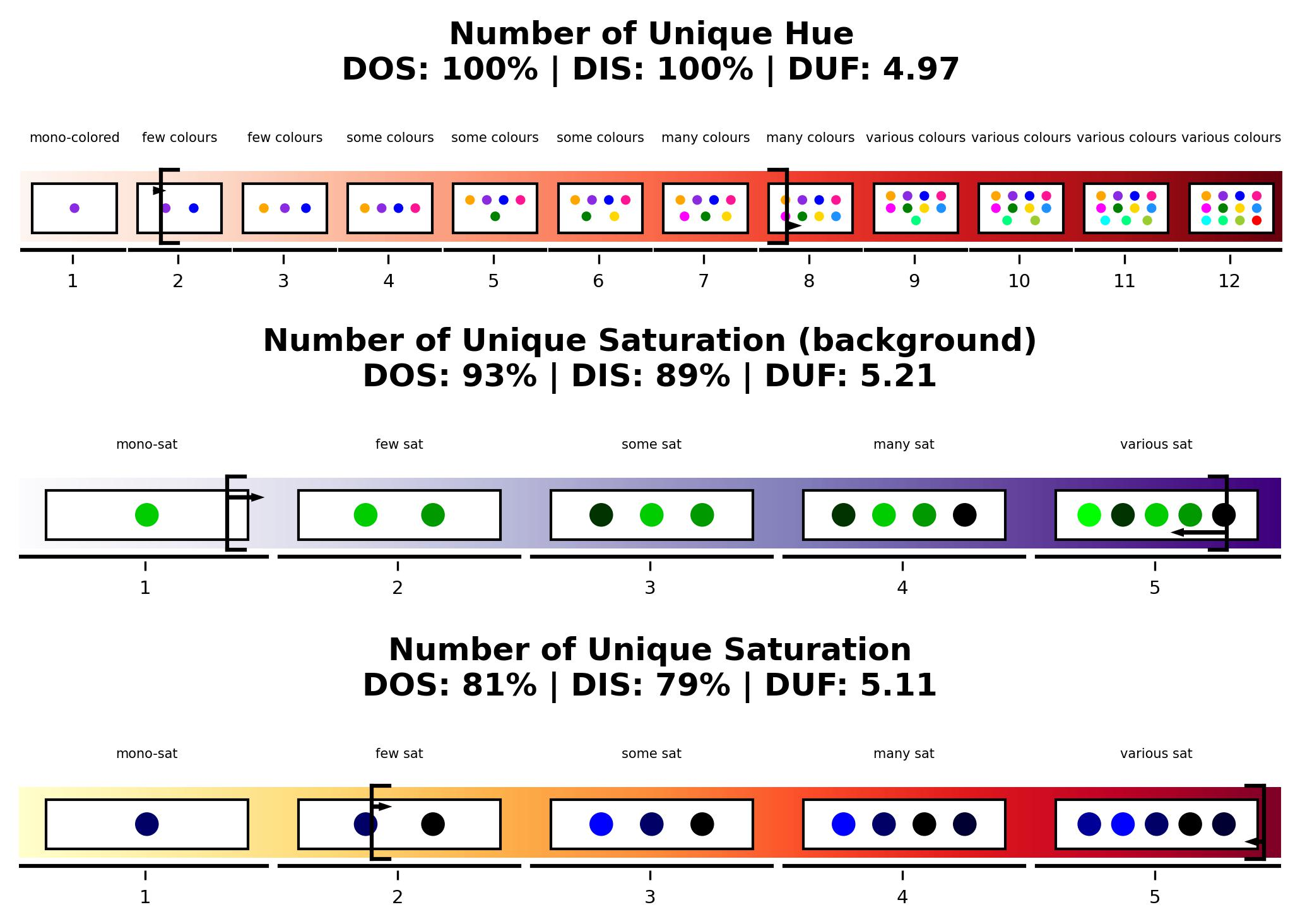}}
    \label{fig:sty}
    \caption{Top-three reasoning results of undetected and misclassified cases.}
    \label{fig:res_top3}
\end{figure*}

The outputs of AI-Reasoner contain 38 charts in total for a reasoning target. Sample outputs are shown in the project's Github\footnote{https://github.com/edgetrier/AI-Reasoner} and in Fig.~\ref{fig:res_top3}. Each chart illustrates the value range (i.e.\ $\DER$) of a \gls{DC} for a defect that was undetected or misclassified by the \gls{IEMRCNN} model. The charts provide the calculated scores (i.e.\ $\DIS$, $\DUF$ and $\DOS$) indicating the importance of a \gls{DC}. These scores are shown at the top of each chart. The reasons for undetected cases include low hue range, neighbouring defects, and low saturation; and the reasons for misclassified cases include low hue range and strong saturation in the background, and narrow hue differences between the inside and outside of the defected area. Based on these findings, the AI-Reasoner suggests the following mitigation strategies: Greyscaling the defects may reduce misclassified but increase the undetected cases. Enhancing the images by normalising the colours of the defects may increase the detected cases and reduce the misclassified cases. 

Zhang \textit{et al.\@}'s \cite{01.IEMRCNN} experiments revealed that using a greyscaled dataset (v2) reduced the misclassified and increased the undetected cases compared to when using the augmented dataset (v1); and when using the image-enhanced dataset (v3), the detected cases were slightly decreased but the misclassified cases were reduced (see Table~\ref{tab:data_dist}). These results are mostly consistent with the mitigation strategies proposed by the AI-Reasoner.

\subsection{Discussion on the suitability of \textit{\gls{SHAP}} for Reasoning with Defect Predictions}
\label{sec:cs1.outcomp}

\begin{figure}[!t] 
    \centering
    \includegraphics[width=0.9\linewidth]{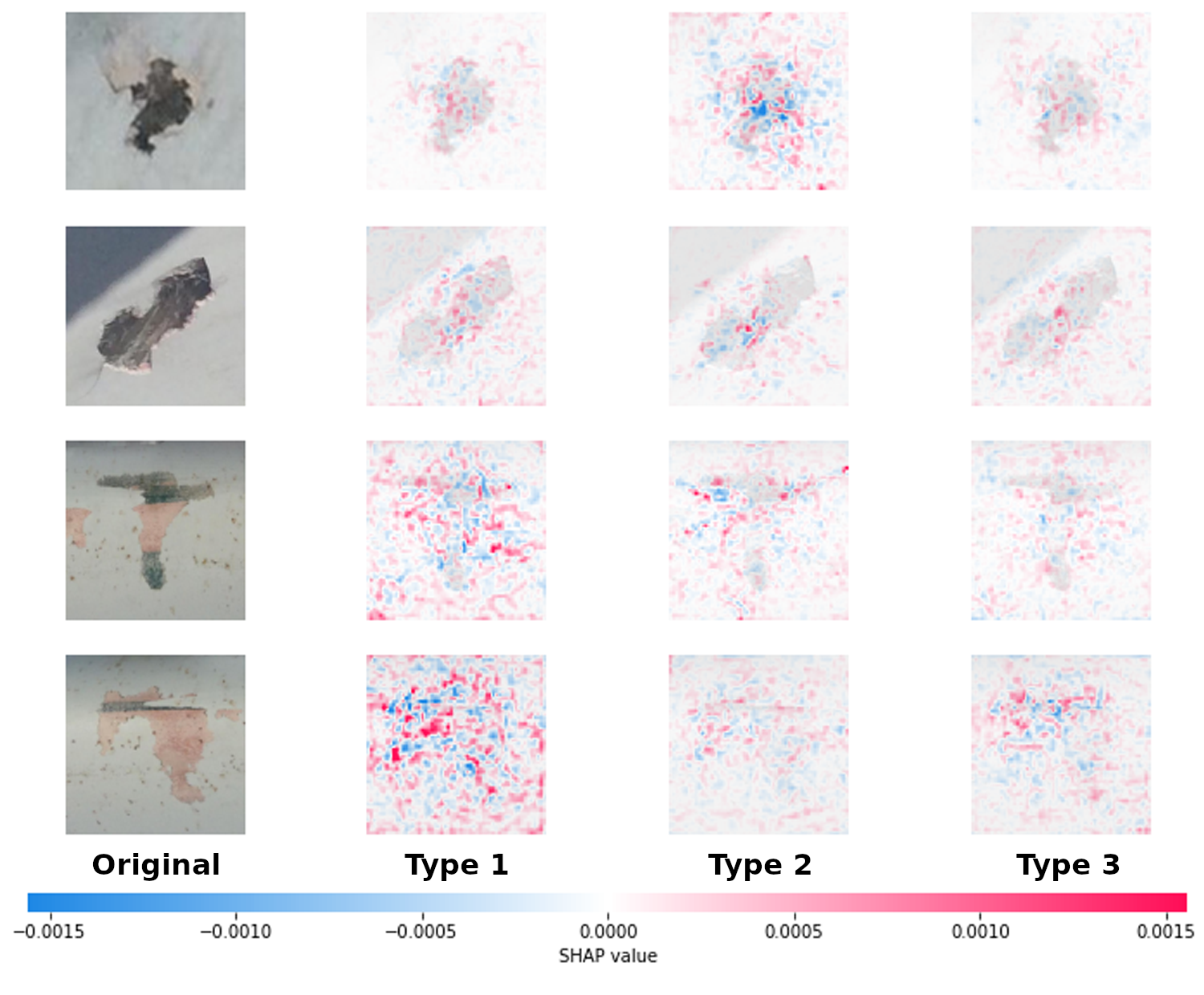}
    \caption{\gls{SHAP} output of four defect images. Important regions are highlighted with blue and red based on the analysis of the hidden layers of the CNN.}
    \label{fig:shapOut}
\end{figure}

This section explains the outputs of a pre-trained CNN model using the \gls{SHAP} algorithm \cite{01.extra_shap}, an \gls{XAI} technique for reasoning with model outputs. \gls{SHAP} cannot support object-detection models and hence a simple CNN model was utilised for the type classification task instead of Zhang et al.'s \cite{01.IEMRCNN} \gls{IEMRCNN} model. Fig.~\ref{fig:shapOut} shows \gls{SHAP}'s output when given four defect images. The important regions are highlighted with blue and red colours based on the analysis of the hidden layers of the CNN. The main observations when comparing \gls{SHAP} with the proposed AI-Reasoner for defect detection, \gls{SHAP} is not compatible with masked-based models that provide multi-channel outputs (i.e.\ matrices that indicate the presence or absence of specific objects or regions within an image) or have complex-structured designs (e.g.\ YOLO, Faster \gls{RCNN}, and Mask \gls{RCNN}). Whereas, the AI-Reasoner is compatible with masked-based models. \gls{SHAP} does not provide details about a defect's characteristics that lead to its misclassification, whereas the AI-Reasoner captures that information in the form of \gls{DC} charts and provides reasoning in the form of charts and textual explanations with mitigation strategies for improving the performance of the model.

\section{Conclusion}
\label{sec:conclusion}
This paper proposes \glspl{DC} and an AI-Reasoner that extracts \glspl{DC} from images and utilises \glspl{DT} to reason with \gls{AI} outputs. The AI-Reasoner provides the end-user with charts, textual explanations and recommendations on improving the dataset pre-processing in order to improve a model's performance. Future work includes experiments with additional datasets, considering other prediction errors (e.g.\ false positive, duplicated predictions, etc.\@), \gls{DC} applications, and exploring the AI-Reasoner's capability in saving experimental time via the mitigation strategies it provides. 

\section*{Acknowledgment} 
\addcontentsline{toc}{section}{Acknowledgment}
The authors acknowledge the expert guidance and datasets provided by Jason Watkins, Chris Gibson, and Andrew Rattray of \textit{Railston \& Co Ltd}.

\bibliographystyle{IEEEtran}
\bibliography{IEEEfull,root}

\end{document}